%% file: main_acl.tex
\pdfoutput=1

\documentclass[11pt]{article}

\usepackage[preprint]{acl}

\usepackage{times}
\usepackage{latexsym}

\usepackage[T1]{fontenc}

\usepackage[utf8]{inputenc}

\usepackage{microtype}

\usepackage{inconsolata}

\usepackage{graphicx}

\usepackage{tcolorbox}
\usepackage{url}
\usepackage{hyperref}
\usepackage{amsmath}
\usepackage{booktabs}
\usepackage{xspace}
\usepackage{pifont}
\usepackage{threeparttable}
\usepackage{tikz}
\usepackage{graphicx} 
\usepackage{pgfplots}
\usepackage{multirow}
\RequirePackage{algorithm}
\RequirePackage{algorithmic}
\RequirePackage{algorithm}
\RequirePackage{algorithmic}
\usepackage{enumitem}
\usepackage{wrapfig}
\usepackage{subcaption}

\usepackage{enumitem}
\usepackage{amsthm}
\usepackage{wrapfig}
\usepackage{xcolor,colortbl}
\usepackage{amssymb}
\usepackage{subcaption}

\newcommand{\byokg}{\textsc{ByoKG-RAG}\xspace}
\newcommand{\glink}{KG-Linker\xspace}

\input{math_commands}

\title{
\byokg: Multi-Strategy Graph Retrieval for Knowledge Graph Question Answering}

\author{
 \textbf{Costas Mavromatis\textsuperscript{1}},
 \textbf{Soji Adeshina\textsuperscript{1}},
 \textbf{Vassilis N. Ioannidis\textsuperscript{1}},
 \textbf{Zhen Han\textsuperscript{1}},
\\
 \textbf{Qi Zhu\textsuperscript{1}},
 \textbf{Ian Robinson\textsuperscript{1}},
 \textbf{Bryan Thompson\textsuperscript{1}},
 \textbf{Huzefa Rangwala\textsuperscript{1}},
 \textbf{George Karypis\textsuperscript{1}}
\\
 \textsuperscript{1}Amazon
\\
 \small{
   \textbf{Correspondence:} \href{mailto:email@domain}{mavrok@amazon.com}
 }
}

\begin{document}
\maketitle
\begin{abstract}
Knowledge graph question answering (KGQA) presents significant challenges due to the structural and semantic variations across input graphs. Existing works rely on Large Language Model (LLM) agents for graph traversal and retrieval; an approach that is sensitive to traversal initialization, as it is prone to entity linking errors and may not generalize well to custom (``bring-your-own'') KGs. We introduce \byokg, a framework that enhances KGQA by synergistically combining LLMs with specialized graph retrieval tools. In \byokg, LLMs generate critical graph artifacts (question entities, candidate answers, reasoning paths, and OpenCypher queries), and graph tools link these artifacts to the KG and retrieve relevant graph context. The retrieved context enables the LLM to iteratively refine its graph linking and retrieval, before final answer generation.
 By retrieving context from different graph tools, \byokg offers a more general and robust solution for QA over custom KGs.
 Through experiments on five benchmarks spanning diverse KG types, we demonstrate that \byokg outperforms the second-best graph retrieval method by 4.5\% points while showing better generalization to custom KGs. \byokg framework is open-sourced at \url{https://github.com/awslabs/graphrag-toolkit}.

\end{abstract}

\section{Introduction}

Knowledge Graphs (KGs)~\citep{vrandevcic2014wikidata} are databases that store information and data in structured format, typically using entities and relationships. The structure of KGs enables efficient data updates and complex queries, making them valuable assets for enterprises across multiple sectors~\citep{ozsoy2024text2cypher}. In the Large Language Model (LLM) era~\citep{brown2020language,bommasani2021opportunities}, KGs have been widely adapted as external information sources to ground LLM responses to semantics present in the graph and to solve KG Question Answering (KGQA) tasks~\citep{pan2024unifying}. 
In  KGQA,  one prevalent approach is retrieval-augmented generation (RAG) that leverages graph retrieval methods to fetch relevant information from the KG as additional input for the LLM~\citep{lewis2020retrieval,peng2024graph}.

\begin{figure}[tb]
    \centering
    \includegraphics[width=\columnwidth]{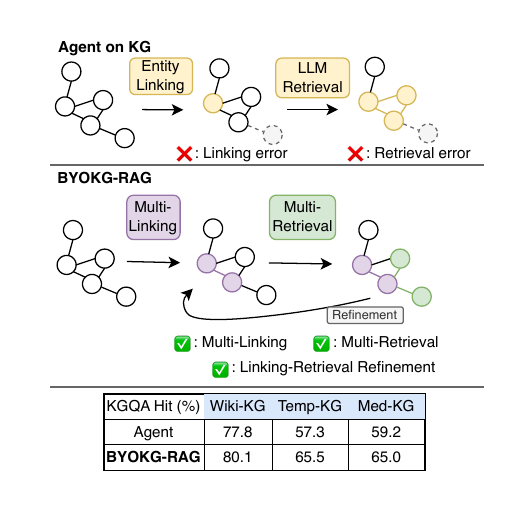}
    \caption{While agentic retrieval (top) may be prone to linking and retrieval errors, e.g., for aggregation queries, \byokg (middle) employs multi-strategy linking and retrieval, along with iterative refinement, to mitigate these errors. As a result, \byokg improves KGQA performance across diverse KGs (bottom). }
    \label{fig:byokg-intro}
    \vspace{-0.2in}

\end{figure}

Adapting RAG to custom (``bring-your-own'')  KGs presents significant challenges as KGs vary not only in their schema design and semantic representations, but also in how information must be retrieved, e.g., via complex aggregation queries.
Current approaches~\citep{mavromatis2024gnn,luo2024rog,jiang2024kg} use graph retrievers specifically fine-tuned for a particular KG, but require training data, which may not be available due to privacy concerns or practical limitations. 
Alternative methods utilize LLMs to traverse the graph step-by-step until arriving at the answers or until termination in an agentic fashion~\citep{jiang2023structgpt,sun2024tog}. These methods leverage the inherent knowledge within LLMs to handle `bring-your-own' scenarios, where training data is unavailable for KGQA. However, such approaches have limitations: Extracting knowledge from custom KGs may require operations like performing complex graph aggregations which is not readily obtained via LLM-based traversal.

\begin{figure*}[tb]
    \centering
    \includegraphics[width=1\linewidth]{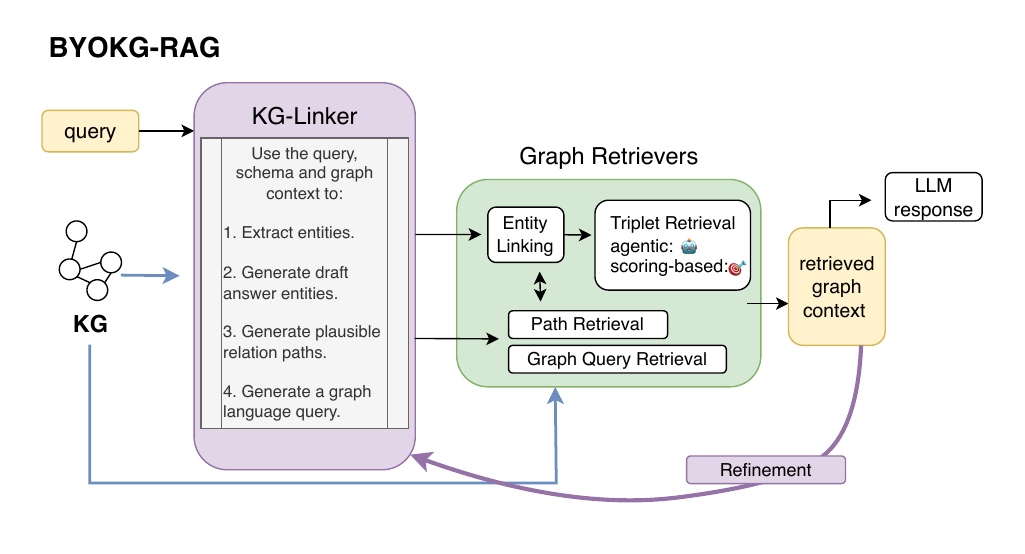}
    \caption{\byokg prompts an LLM to generate critical graph artifacts for graph linking. Then, a toolkit of graph retrievers operates on the underlying graph and, based on the generated artifacts and query's context, retrieves relevant information for final KGQA or linking refinement.}
    \label{fig:byokg-rag}
\end{figure*}

In this work, we introduce \textbf{\byokg}, a framework that addresses these challenges through multi-strategy graph retrieval. Our key insight is to leverage LLMs as a \glink that generates diverse graph artifacts - from entity mentions to executable queries. These artifacts are then processed by specialized graph tools, each designed to handle different retrieval scenarios effectively.
Given a user query and the  KG schema\footnote{The schema typically expresses the node and relation types present in the graph, among other properties, and is available in common graph databases via \texttt{graph.get\_schema()} functionality.}, the LLM is prompted to generate relevant graph artifacts, such as question and answer entities, relationship types, and executable graph queries. Subsequently, the graph toolkit takes these artifacts as input, links them to the KG and retrieves relevant graph context through  different retrieval methods. The retrieved context can be used in an iterative fashion so that \glink improves the accuracy of both the generated artifacts and subsequent retrieval, before final KGQA.
Compared to agentic retrieval, which is restricted to the graph context that can be explored by the agent, \byokg retrieves context from multiple linking and retrieval tools, including agentic retrieval, to enhance KGQA accuracy; see \Figref{fig:byokg-intro} for a high-level comparison.  

Our contributions are summarized below:
\begin{itemize}
    \item We introduce \byokg, a novel framework that improves KGQA across custom KGs via multi-strategy graph linking and retrieval. 
    \item \byokg outperforms the best graph retrieval method by 4.5\% points in KGQA across four KGs. In addition, \byokg achieves on-par or better performance than state-of-the-art KG agents on key benchmarks. 
    \item Our thorough evaluation across KGs with diverse semantics demonstrates the benefits of multi-strategy graph retrieval, and we open-source our toolkit.
    
\end{itemize}

\section{Background \& Related Work}

\subsection{Problem Formulation}

Consider a Knowledge Graph (KG) $\gG$ containing facts represented as triplets $(h,r,t)$, where $h$ represents the head entity, $t$ represents the tail entity, and $r$ denotes the relation between them. The task of KGQA involves extracting a set of entities $\gA \in \gG$ that correctly answer a given natural language question $q$. Since KGs typically contain millions of facts and nodes, retrieval-augmented generation  retrieves relevant graph context $\gC$, which is then used by LLM to generate the answers as $\gA = \text{LLM}(q, \gC)$. To retrieve $\gC$, common approaches combine entity linking techniques~\citep{yih2015semantic} to identify question entities ${e_q}$ with graph retrieval methods that extract relevant information within an $L$-hop neighborhood of these entities~\citep{sun-etal-2018-open}.

\subsection{Related Works}
\textbf{KGQA Methods}. 
KGQA methods involve either fine-tuning or zero-shot/few-shot inference. Fine-tuning methods encompass several approaches. Some methods focus on parsing the given question into a logical form query executable over the KG ~\citep{sun2020sparqa,lan-jiang-2020-query,ye-etal-2022-rng,gu-etal-2023-dont,agarwal2023bring}. Others involve training specialized neural networks for handling KG data~\citep{sun-etal-2018-open,he2021improving, mavromatis2022rearev,jiang2023unikgqa}. Recent approaches aim at teaching LLMs to better understand KG semantics~\citep{luo2024rog,ao2025lightprof} or instructing the LLM on how to explore the graph~\citep{jiang2024kg,luo2025kbqao1,zhang2025learning}. 
Zero-shot and few-shot methods typically  rely on LLMs' knowledge and combine them with appropriate graph tools for graph traversal~\citep{jiang2023structgpt, kim2023kggpt, sun2024tog,dong-2025-effiqa,sui2024fidelis,jo2025r2}, graph querying~\citep{li2023binder,wang2023knowledge}, and constrained decoding on KGs~\citep{luo2024graph,li2024decoding}. 
\byokg is a novel framework for improving graph linking and retrieval in ``bring-your-own-KG'' (zero-shot) settings.

\textbf{Graph RAG}. Graph RAG refers to the framework of using context originating from a graph or a KG within RAG~\citep{edge2024local,he2024gretriever,mavromatis2024gnn}. Recently, methods~\citep{gutierrez2024hipporag,guo2024lightrag,luo2025gfm,chen2025pathrag} use LLMs for constructing KGs out of raw data to improve information retrieval  with graph information ~\citep{lee2024hybgrag,sarmah2024hybridrag}, beyond text-based semantics. With the increased deployment of KG-powered applications~\citep{peng2024graph}, \byokg offers a multi-strategy toolkit for retrieving task-specific graph information.

\textbf{Graph Linking}. 
Entity linking~\citep{yadav2019survey} in KGs maps mentions in natural language to their corresponding graph entities and has been extensively studied in KGQA systems~\citep{yih2015semantic,oliya2021end,li2020efficient}. 
Beyond entity linking, some approaches focus on mapping natural language questions to graph paths~\citep{saxena2020improving,shi2021transfernet} and subgraphs via graph query execution~\citep{lan2020query,gu2022arcaneqa,shu2022tiara,yu2022decaf,gao2025promoting}, typically employing training data for more accurate linking and retrieval.
\glink extends these ideas by combining multiple linking approaches with LLM capabilities, enabling robust graph retrieval without task-specific training data.

\section{Bring-Your-Own-KG RAG (\byokg)}
\label{sec:kglink}

In \byokg, we address the challenges of graph retrieval in custom KGs through a novel two-stage approach. First, we prompt an LLM (\glink) to generate graph artifacts - including entities, paths, and queries - that serve as anchor points for graph retrieval. Then, these artifacts are processed by specialized graph tools that link them to the underlying KG and retrieve relevant context through multiple complementary methods. This synergistic combination of LLM-generated artifacts and specialized graph tools enables more robust retrieval across custom KGs. 
The overall framework is  illustrated in \Figref{fig:byokg-rag}.

\subsection{\glink}
The key insight behind \glink is to leverage LLMs' natural language understanding to generate diverse graph artifacts that can be linked to the KG through specialized tools.
In \glink, given an input a user query, a graph schema, and an optional graph context, an LLM is  prompted  to generate graph artifacts, including question entities, draft answers, useful relation paths, and graph queries. The LLM's prompt includes four main tasks: (1) Entity Extraction,  (2) Relation Path Identification, and (3) Graph Query Generation, and (4): Draft Answer Generation.  

The instruction template $p$ is presented in \Figref{fig:gconnector-prompt}, where \texttt{\{question\}} is the input user query, \texttt{\{schema\}} is the KG schema, including node types and relations, and \texttt{\{graph\_context\}} denotes the graph context retrieved from intermediate steps (if available). \texttt{\{openCypher\}} may be replaced with other graph query languages.
Based on the LLM prompt, the generated graph artifacts are the following,
\begin{align}
    \tilde{\gE} &= \text{LLM(task = ``Entity Extraction'')}, \\
    \tilde{\gP} &= \text{LLM(task = ``Path Identification'')}, \\
    \tilde{\gQ} &= \text{LLM(task = ``Graph Query Generation'')}, \\
    \tilde{\gA} &= \text{LLM(task = ``Draft Answering'')},
\end{align}
where $\tilde{\gE}$, $ \tilde{\gP}$, $\tilde{\gQ}$, and $\tilde{\gA}$ denote the generated entities, paths, graph query, and answer entities, respectively. Note that those graph artifacts can be generated by a single LLM call.

\input{figures/llm_connector}

\subsection{Entity Linking}
\label{sec:linking}
Entity linking is fundamental to our framework, enabling graph traversal from accurate graph anchor points. Entity linking maps LLM-generated entities $\hat{\gE} = \tilde{\gE}  \cup \tilde{\gA}$ to actual entities $\gE$ in the graph, e.g., ``\textit{Jamaican people}'' $\mapsto$ ``\textit{Jamaica}''.
We use the textual descriptions of the entities to perform entity linking and implement two complementary linking methods: fuzzy-string matching, and node embedding linking. 

For each entity $\hat{e}_i \in \hat{\gE}$, we retrieve top-$m$ entities present in the graph based on string match scores, node embedding similarity, or the union of both, where $m$ is a hyperparameter, defaulting to $m=3$. By default, we use string-based matching$\footnote{https://github.com/seatgeek/thefuzz}$ and the encoder used for transforming entity names into embeddings is \texttt{bge-m3}~\citep{chen2024bge}. After performing entity linking, we obtain the linked entity set $\gE$, where $|\gE| \leq m \cdot |\hat{\gE}|$  when using single linking method, or $|\gE| \leq 2 m \cdot |\hat{\gE}|$ when using both string and embedding-based linking.

\subsection{Path Retrieval}
Path retrieval executes and validates the LLM-generated paths while discovering alternative routes in the graph that can lead to the required information.
Formally, it takes as input the generated chain of relations $\tilde{\gP}$, executes them on the graph, and returns the resulting intermediate entities along with the relation paths. We implement a \texttt{follow-paths} functionality which takes as input a sequence of relations $\tilde{\gP}$ and source entities $\gE$ and returns the executable paths $\gP_f \subset \tilde{\gP}$ over the graph $\gG$ as
\begin{equation}
    \gP_f = \text{Follow-Paths}(\tilde{\gP}, \gE, \gG).
\end{equation}
We perform a breadth-first search starting from the source entities, exploring the graph until all valid paths generated by KG-Linker are found or no further valid paths remain. As source entities, we use the linked entities $\gE$ of the entity linking step~\Secref{sec:linking}.

In addition, we retrieve the shortest paths connecting the extracted entities $\gE$ and candidate answers $\gA$ generated by the LLM (if non-empty). We implement a \texttt{shortest-paths} functionality which returns
\begin{equation}
    \gP_s = \text{Shortest-Paths}(\gE, \gA, \gG)
\end{equation}
based on the candidate entities and answers, using the Dijkstra algorithm\footnote{https://en.wikipedia.org/wiki/Dijkstra\%27s\_algorithm}. Path retrieval returns
\begin{equation}
    \gP = \gP_f \cup \gP_s.
\end{equation}

\subsection{Graph Query Retrieval}
Graph query retrieval translates natural language into executable graph operations, crucial for handling complex queries in enterprise KGs.
As a default, we prompt the LLM to generate a query $\tilde{Q}$ in openCypher language which is common in graph databases like Neptune and Neo4j. This step requires the graph to be stored in a format that supports the execution of the generated graph language queries. We implement  an \texttt{execute-query} functionality which takes as input a query $\tilde{Q}$ and returns 
\begin{equation}
    \gQ_a = \text{Execute-Query}(\tilde{Q}, \gG),
\end{equation}
which are the query execution results over $\gG$.

\subsection{Triplet  Retrieval}
\label{sec:agentic}
Triplet retrieval complements the above methods by capturing relevant facts that might be missed through direct retrieval. Triplet retrieval operates on the triplet-level granularity of the KG, where triplets $\gT$ are formatted as $(h,r,t)$ and express natural language descriptions as ``\textit{head $\xrightarrow[]{}$ relation $\xrightarrow[]{}$ tail}'', e.g., ``\textit{Jamaica →
language\_spoken → English}''. We implement two approaches: agentic retrieval for step-by-step exploration, and scoring-based retrieval for more efficient semantic matching.

\input{algorithms/agentic}
\input{algorithms/refine}

\textbf{Agentic Retrieval.}
Given the linked entities $\gE$ and the query $q$, agentic  retrieval  fetches the relevant triplets $\gT_q$, using an LLM that iteratively traverses the KG based on $q$. We provide the algorithm in Algorithm~\ref{alg:agentic} and the LLM prompts are in Appendix~\ref{app:agent-prompt}.

At each iteration $t$, we obtain one-hop candidate triplets $\gT_q^t$ and prompt the LLM to  select  the most relevant relations $\gR_q^t$. The triplets are filtered based on the LLM's selection and subsequently we prompt the LLM to select the most relevant entities  $\gE_q^t$ to explore next. The process is terminated until self-termination (empty entity set to explore) or until we reach the maximum agentic iterations $T_A$, defaulting to $T_A=3$. The agent returns the graph context $\gT_q$ explored until termination.

\textbf{Scoring-based  Retrieval.}
Alternative to LLM-based traversal, scoring-based retrieval performs top-$k$ triplet selection $\gT_q$ based on question-triplet semantic similarity, where $|\gT_q| = k$ and $k$ is a hyperparameter. In Appendix~\ref{app:scoring}, we present an implementation that uses reranker models to select relevant triplets within $L$-hop distance from linked entities. Our framework  also supports a more efficient embedding-based retrieval approach.

\subsection{Iterative Process}
\label{sec:refine}
\byokg's iterative process enables progressive refinement of retrieved context. Each iteration combines information from multiple retrieval methods, allowing the LLM to generate more accurate artifacts based on accumulated context.
\byokg's iterative process is presented Algorithm~\ref{alg:refine}.  After graph linking and retrieval, we collect the retrieved context 
\begin{equation}
    \gC = \gT_q \cup \gP_f \cup \gP_s \cup \gQ_a,
\end{equation}
and verbalize it with predefined templates. As shown in Algorithm~\ref{alg:refine}, the context $\gC$ is used as input to \glink's prompt (\Figref{fig:gconnector-prompt}) to iteratively refine its generations $\tilde{\gE}$, $ \tilde{\gP}$, $\tilde{\gQ}$, and $\tilde{\gA}$.  
By default, we have $T_R = 2$ iterations, but we include self-termination when the \glink does not provide new entities to link. Finally, we collect the graph context $\gC$ as input for the final KGQA task.

\section{Experimental Setup}

\subsection{Datasets \& Metrics} 

We evaluate KGQA benchmarks spanning different backbone KGs. The statistics are provided in Appendix~\ref{app:data}.

\textbf{WebQSP}~\citep{yih2015semantic} and ComplexWebQuestions (\textbf{CWQ})~\citep{talmor2018web} involve questions answerable over the Freebase KG~\citep{bollacker2008freebase}. WebQSP and CWQ contain 1,628 and 3,531 questions, respectively, and involve general knowledge about entities, requiring 1-2 hop and 1-4 hop KG traversal, respectively. WebQSP-IH and CWQ-IH refer to in-house subsets, containing 500 questions each from the original set. For pre-processing the Freebase KG, we follow standard steps from previous works~\citep{he2021improving}.

\textbf{CronQuestions}~\citep{saxena2021question} are questions that require temporal reasoning over a KG subset of Wikidata, which contains temporal information~\citep{lacroix2020tensor}. As the tempral KG contains quadruples (\textit{head}, \textit{relation}, \textit{tail}, 
 \textit{timestamp}), we convert them to triplets (\textit{head}, \textit{relation: timestamp}, \textit{tail}).  We focus on CronQuestions's subset of complex questions and we sample 200 per question type (before/after, first/last, time join) resulting into 600 questions. 

\textbf{MedQA}~\citep{jin2021disease} consists of questions extracted from USMLE exams. We combine MedQA with Disease DrugBank KG~\citep{wishart2018drugbank}, following previous works~\citep{yasunaga2021qagnn}. Since the KG may not be relevant for all questions, we filter for questions whose answer candidates appear in the KG (via entity matching). Additionally, we remove MedQA's multiple choices from the LLM's context to increase the benchmark's difficulty.

The \textbf{Text2cypher} dataset~\citep{ozsoy2024text2cypher} emulates text2cypher queries encountered in enterprise KGs, such as Northwind. As not all ground-truth queries are executable over the KGs provided in the benchmark, we filter for questions whose ground-truth answers can be retrieved from the graphs. %

\input{tables/main_results}

\textbf{Metrics.} Similar to previous works, we report the \textit{Hit} metric for WebQSP, CWQ, and CronQuestions, which measures if the LLM generates any correct answer via exact string matching. In MedQA, we report Hit@2 (\textit{H@2}) by augmenting the LLM generation based on the retrieved KG context with its original prediction, and measure if correct answers are found. For Text2cypher, we use Claude-Sonnet-3.5 as an LLM-as-a-judge (\textit{LLMaaJ}) that scores if the executed results match the ground-truth results (1.0: correct, 0.5: partially correct, 0.0: incorrect). We also evaluate retrieval accuracy by Recall@$k$ which measures if correct answers are found within top-$k$ retrieval results.

\subsection{Competing Methods}

\textbf{Baselines.}  As key baselines, we include components present within \byokg's framework. \textit{Vanilla LLM} is the approach of direct QA without KG context. \textit{LLM+graph-query} prompts the LLM to generate an executable graph query, and augments the LLM's direct predictions along with the execution results. For MedQA, we prompt the LLM to generate answers based on the execution results. \textit{Text-based Retrieval} (Appendix~\ref{app:text-retrieval}) retrieves top-$k$ triplets based on embedding similarity, while \textit{Graph Reranker} (Appendix~\ref{app:graph-reranker}) uses a reranker to select top-$k$ triplets within $L=2$ hops of linked entities. We use $k=50$ for Freebase, $k=10$ for the rest KGs.  \textit{Agentic Traversal} (Algorithm~\ref{alg:agentic}) is the LLM-based KG traversal and retrieval as described in Algorithm~\ref{alg:agentic}. Note that Graph Reranker and Agentic Traversal require entity linking as a first step, and we also incorporate entity linking in LLM+graph-query for better query generation.  We employ the same LLM across methods for downstream KGQA, after graph retrieval. 

For a fair comparison, we focus on zero/few-shot methods and evaluate against established KG agents, including those using agentic traversal~\citep{jiang2023structgpt,sun2024tog,sui2024fidelis} and LLM-guided graph retrieval~\citep{li2023binder,dong-2025-effiqa}.

\textbf{\byokg Implementation.} We implement \byokg with default hyperparameters as described within subsections of \Secref{sec:kglink}. We use entity linking with top-$m=3$ via combining string and embedding-based matching. When executing graph queries, we store the KG in NeptuneAnalytics\footnote{https://aws.amazon.com/neptune/} and query the database with openCypher\footnote{https://opencypher.org/}. We retrieve triplets via agentic-based retrieval, and we combine it with top-$k=10$ efficient text-based retrieval. We also present results for `\byokg (scoring)', which substitutes the agentic retrieval with the scoring-based approach in Appendix~\ref{app:graph-reranker}. For the LLMs used in \glink, we experiment with Claude-Sonnet-3.5, Claude-Haiku-3.5~\citep{claude}, and Llama-3.3-70B~\citep{grattafiori2024llama}, accessed via Bedrock API\footnote{https://aws.amazon.com/bedrock/}.

\section{Experimental Results}

\subsection{Main Results}

Table~\ref{tab:main-results} demonstrates the effectiveness of \byokg across diverse KGQA benchmarks. On the Freebase-KG benchmarks, \byokg achieves 86.6\% and 73.6\% Hit rates on WebQSP-IH and CWQ-IH respectively, outperforming the best baseline methods (86.2\% and 69.3\%). The improvement is also pronounced on specialized KGs: for temporal reasoning on CronQuestions, \byokg achieves a 65.5\% Hit rate, showing a significant gain over text-based retrieval (59.8\%). Similarly, on the medical domain MedQA benchmark, \byokg improves H@2 performance to 65.0\% compared to graph-query's 59.2\%. The framework's generalization capability is further demonstrated on enterprise KGs, where it achieves 64.9\% LLMaaJ score on Northwind, outperforming graph-query (55.3\%).
Table~\ref{tab:main-results} also evaluates \byokg  using different LLM backbones. While Claude-Sonnet-3.5 performs best overall, the performance improvements are consistent across Claude-Haiku-3.5 and Llama-3.3-70B, indicating that our framework's benefits are robust across different LLMs. Using Claude-Sonnet-3.5, \byokg outperforms the strongest baseline across benchmarks (agentic retrieval for WebQSP-IH/CWQ-IH, text-based retrieval for CronQuestions, and graph querying for MedQA/Northwind) by \textbf{4.5\% points}, on average.

\input{tables/freebase_oracle}

In Table~\ref{tab:freebase-results}, we compare \byokg against state-of-the-art zero/few-shot KGQA methods on WebQSP and CWQ benchmarks. \byokg achieves the best performance on WebQSP (87.1\% Hit rate) and comparable results on CWQ (71.1\%), while requiring fewer LLM calls than competing methods. Notably,  \byokg's more efficient scoring-based variant maintains strong performance while using only 2-3 LLM calls.

\subsection{Ablation Studies}

Our studies in Tables~\ref{tab:main-results},~\ref{tab:linker-results},~\ref{tab:retrieval},  and~\ref{tab:latency-results} analyze \byokg's components.

Table~\ref{tab:main-results} reveals the effectiveness of different retrieval methods in \byokg across diverse KG types. On Freebase benchmarks (WebQSP-IH, CWQ-IH), agentic traversal shows strong performance (86.2\%, 69.3\% with Claude-Sonnet-3.5), indicating step-by-step graph exploration is well-suited for multi-hop queries. For temporal reasoning in CronQuestions, text-based retrieval proves effective (60.2\% Hit  with Llama-3.3-70B), suggesting temporal information can be captured through semantic matching. For the medical domain (MedQA), the combination of different retrieval methods in \byokg yields notable improvements (65.0\% H@2 with Claude-Sonnet-3.5). Most distinctively, on enterprise KGs (Northwind), the ability to generate and execute Cypher queries proves crucial, with \byokg achieving 64.9\% accuracy and LLM+graph-query achieving 55.3\% with Claude-Sonnet-3.5.

\input{tables/ablation}

\input{tables/entity_linking}

Table~\ref{tab:linker-results} demonstrates the value of multi-strategy linking beyond entity matching alone. On CWQ-IH, \glink improves performance from 63.0\% to 71.6\% with scoring-based retrieval and from 69.3\% to 73.6\% with agentic retrieval. Similar gains are observed on CronQuestions. 
\input{tables/case}

Table~\ref{tab:retrieval} compares retrieval effectiveness across different methods. While supervised methods like RoG, SubgraphRAG, and GNN-RAG achieve Recall@10 scores of 54.5-64.1\% using substantial training data (27.6K-30.5K examples), \byokg performs competitively without any training data. \byokg's agentic variant achieves 70.5\% Recall@10 while retrieving 396 tokens in total, while its scoring-based variant achieves 60.6\% with 1,483 tokens, demonstrating effective zero-shot retrieval with moderate context sizes.

Furthermore, we provide a case study on how \byokg iteratively improves its retrieval in Table~\ref{tab:case}. 

Additional experiments and cases studies are in Appendix~\ref{app:exps} and~\ref{app:case}.

\section{Conclusion}

We introduced \byokg, a framework that enhances KGQA by combining LLMs with graph retrieval tools. Through extensive experiments across five benchmarks, we demonstrated that multi-strategy graph retrieval matters: \byokg leverages multiple retrieval methods to achieve superior performance (4.5 percentage points improvement over the strongest baselines) while generalizing effectively to KGs with diverse semantics.

\section*{Limitations}

\byokg leverages diverse graph tools to retrieve context, enabling more effective KGQA across custom KGs. However, without proper context pruning mechanisms, the retrieved context may become too lengthy and potentially confuse LLMs with limited context-handling capabilities~\citep{liu2023lost}. Furthermore, our current work considers the KG as the sole source of external information. Future work could expand \byokg's capabilities by incorporating additional information sources, such as text databases~\citep{wu2024stark}, thereby enhancing the framework's applications.

\bibliography{KGQA}

\input{appendix}

\end{document}

%% file: math_commands.tex
\usepackage{amsmath,amsfonts,bm}

\def\Figref#1{Figure~\ref{#1}}

\def\Secref#1{Section~\ref{#1}}

\def\eqref#1{equation~\ref{#1}}

\def\1{\bm{1}}

\DeclareMathAlphabet{\mathsfit}{\encodingdefault}{\sfdefault}{m}{sl}
\SetMathAlphabet{\mathsfit}{bold}{\encodingdefault}{\sfdefault}{bx}{n}

\def\gA{{\mathcal{A}}}

\def\gC{{\mathcal{C}}}

\def\gE{{\mathcal{E}}}

\def\gG{{\mathcal{G}}}

\def\gP{{\mathcal{P}}}
\def\gQ{{\mathcal{Q}}}
\def\gR{{\mathcal{R}}}

\def\gT{{\mathcal{T}}}

%% file: figures/llm_connector.tex
\begin{figure}
\begin{minipage}{1\columnwidth}
    \centering
\begin{tcolorbox}[title=\glink's Prompt $p$.]
        \small
        Given a question, schema, and optional graph context, your role is to perform the following tasks:
        
        \vspace{10pt}
        \textbf{Task: Entity Extraction} \\
        Extract all topic entities from the question and useful intermediate entities from the graph context (if provided) within  \texttt{<entities>} tags.

        \vspace{10pt}
        \textbf{Task: Relationship Path Identification} \\
        Identify all relevant relationship paths that connect the entities and can be used to answer the question within \texttt{<paths>} tags.

        \vspace{10pt}
        \textbf{Task: Graph Query Generation} \\
        Construct a complete, executable graph query statement that will retrieve the answer from a graph database. Format your query in \{openCypher\} language within \texttt{<opencypher>} tags.

        \vspace{10pt}
        \textbf{Task: Draft Answer Generation} \\
        Answer the question using your existing knowledge base or the external information  in the graph context (if provided) within \texttt{<answer>} tags.
        
        \vspace{10pt}
        Now, please analyze the following:
        
        \vspace{5pt}
        \textbf{Question}: \{question\} \\
        \textbf{Schema}: \{schema\} \\
        \textbf{Graph Context}: \{graph\_context\}
    \end{tcolorbox}

\caption{The prompt template used in \glink. Full prompts are provided in Appendix~\ref{app:linker-prompt}.}
\label{fig:gconnector-prompt}
\end{minipage}
\end{figure}

%% file: algorithms/agentic.tex
\begin{algorithm}[t]
\centering
\caption{Agentic Retrieval.} \label{alg:agentic}
\begin{algorithmic}[1]
   \STATE {\bfseries Input:} Query  $q$, Entities $\gE$, Graph $\gG$, Iterations $T_A$
   \STATE {\bfseries Optional:} Context $\gT_q^{0}$, else $\gT_q^{0} = \emptyset$
   \STATE $\gE_q^{0} \gets \gE$
   
   \FORALL{$t \in T_A$} 
   
    \STATE $\hat{\gT}_q^{t} =  \text{OneHop}(\gE_q^{t-1},\gG)$ \hfill\COMMENT{\textit{one-hop graph expansion}}
    \STATE $\gR_q^{t} = \text{LLM}(\textit{``Select Relations''}, \hat{\gT}_q^{t}, q)$ \hfill\COMMENT{\textit{relevant relations}}
    \STATE $\hat{\gT}_q^{t} \gets \text{Filter}(\hat{\gT}_q^{t}: r \in \gR_q^{t})$
    \STATE $\gT_q^{t} \gets \gT_q^{t-1} \cup \hat{\gT}_q^{t} $ \hfill\COMMENT{\textit{updated context}}
    \STATE $\gE_q^{t} = \text{LLM}(\textit{``Select Entities''}, \gT_q^{t}, q)$
    \hfill\COMMENT{\textit{next entities}}
    \IF{$\gE_q^{t} == \emptyset$}
        \STATE {\bfseries break} \hfill\COMMENT{\textit{self-termination}}
    \ENDIF
   \ENDFOR
   \STATE  $\gT_q \gets  \gT_q^{t}$.
   \STATE {\bfseries Output:} Return $\gT_q$.
\end{algorithmic}
\end{algorithm}

%% file: algorithms/refine.tex
\begin{algorithm}[t]
\centering
\caption{\byokg Framework.} \label{alg:refine}
\begin{algorithmic}[1]
   \STATE {\bfseries Input:} Prompt $p$, Query  $q$, Schema $S$, Iterations $T_R$
   \STATE $\gC \gets \emptyset$
   \FORALL{$t \in T_R$} 
   \STATE $\tilde{\gE}, \tilde{\gP}, \tilde{\gQ}, \tilde{\gA} = $ LLM($p$, $q$, $S$, $\gC$)
   \STATE $\gE, \gA = $ EntityLink($\tilde{\gE}, \tilde{\gA}$)
   \STATE {\bfseries if} $\tilde{\gE} = \emptyset$: {\bfseries break}
   \STATE $\gP = $ PathRetrieve($\tilde{\gP}, \tilde{\gE}, \tilde{\gA}$)
   \STATE $\gQ_a$ = QueryRetrieve($\tilde{\gQ}$)
   \STATE $\gT_q$ = TripletRetrieve($q, \gE \cup \gA, \gC$)
   \STATE $\gC \gets \gT_q \cup \gP \cup \gQ_a \cup \gC$
    
   \ENDFOR
   \STATE  $\gA  =$ LLM($q, \gC$).
   \STATE {\bfseries Output:} Answers $\gA$.
\end{algorithmic}
\end{algorithm}

%% file: tables/main_results.tex
\begin{table*}[tb]
	\centering
	\caption{Performance comparison of zero-shot  methods across different KGQA benchmarks. We  bold the best and underline the second-best method.}
	\label{tab:main-results}%
	\resizebox{0.8\linewidth}{!}{
	\begin{threeparttable}
		\begin{tabular}{@{}l|ccccc@{}}
			\toprule
          &  \multicolumn{2}{c}{Freebase-KG} & Temp-Wikidata & DiseaseDB-KG  & \multicolumn{1}{c}{Cypher-KG}\\
         &  WebQSP-IH & CWQ-IH & CronQuestions & MedQA & Northwind \\
         &  Hit (\%) & Hit (\%)  & Hit (\%) & H@2 (\%) & LLMaaJ (\%)  \\
         \midrule
         \midrule
         &  \multicolumn{5}{c}{\textit{Claude-Sonnet-3.5}} \\
        Vanilla LLM &  70.4 & 54.2 & 19.2 & 57.9 & 0.0 \\
         LLM + graph-query$^\alpha$& 75.2 & 54.3 & 19.2 & \underline{62.5} & \underline{55.3}  \\
        
        Text-based Retrieval$^\beta$ & 70.8 & 56.8 & \underline{59.8} & 58.1 & 0.7\\
        Graph Reranker$^\gamma$ & 76.0 & 63.0 & 41.8 & 58.2 & 3.4\\
         Agentic Traversal$^\delta$ & \underline{86.2} & \underline{69.3} & 57.3 & 59.2 & 3.4  \\
        \midrule
            \textbf{\byokg}  & \textbf{86.6}  & \textbf{73.6 }& \textbf{65.5}  & \textbf{65.0}  & \textbf{64.9} \\

        \midrule
        \midrule
        &  \multicolumn{5}{c}{\textit{Claude-Haiku-3.5}} \\
        Vanilla LLM & 67.2 & 48.2 &  15.0 & 44.7 & 0.0\\
        LLM + graph-query$^\alpha$& 76.4 & 48.4 & 15.0 & \underline{49.8} & \underline{52.6}\\
        Text-based Retrieval$^\beta$  & 74.6 & 58.4 & \underline{59.6} & 47.6 &  0.0\\
        Graph Reranker$^\gamma$ & 75.6 & 62.7 & 38.7 & 46.6 & 3.4 \\
        Agentic Traversal$^\delta$ & \underline{81.0} & \underline{64.2} & 40.8 & 46.2  & 3.4 \\
        \midrule
        \textbf{\byokg} & \textbf{82.8} & \textbf{66.8} & \textbf{61.9} & \textbf{54.2} & \textbf{59.1}\\
        \midrule
        \midrule
         &  \multicolumn{5}{c}{\textit{Llama-3.3-70B}} \\
         Vanilla LLM & 68.2 & 51.8 & 14.8 & 45.1 & 0.0 \\
         LLM + graph-query$^\alpha$& 69.2 & 51.8 & 14.8 & \underline{49.1}  & \underline{58.6} \\
         Text-based Retrieval$^\beta$ & 73.2 & 60.1 & \underline{60.2} & 47.3& 0.0 \\
         Graph Reranker$^\gamma$ & 75.6 & 62.9 & 38.9 & 46.5& 3.4 \\
        Agentic Traversal$^\delta$ & \underline{81.9}  & \underline{67.0} & 45.8 & 45.1 & 3.4 \\
        \midrule
        \textbf{\byokg} & \textbf{82.4} & \textbf{67.2} & \textbf{62.2} & \textbf{53.1} & \textbf{68.9}\\

			\bottomrule
		\end{tabular}%
            \small
            \begin{tablenotes}
            \item $^\alpha$We combine LLM generation with text2cypher generation. 
            \item $^\beta$We retrieve top-$k$ triplets based on question-triplet embedding similarity (Appendix~\ref{app:text-retrieval}).
            \item $^\gamma$We retrieve top-$k$ triplets via graph-based reranking (Appendix~\ref{app:graph-reranker}).
            \item $^\delta$The LLM agent iteratively explores relevant entities and relations until self-termination (Algorithm~\ref{alg:agentic}).
            \end{tablenotes}
		\end{threeparttable}
}

\end{table*}%

%% file: tables/freebase_oracle.tex
\begin{table}[tb]
	\centering
        	
        \caption{Performance comparison of \byokg with state-of-the-art zero/few-shot KGQA methods on WebQSP and CWQ datasets. We provide performance numbers of competing methods  as reported in the corresponding literature. In addition, we report average number of LLM calls as a metric  in terms of LLM utilization (the lower, the better).}
        \label{tab:freebase-results}%
	\resizebox{\columnwidth}{!}{
    	
	\begin{threeparttable}
		\begin{tabular}{@{}l|cc|cc@{}}
			\toprule
			  & \multicolumn{4}{c}{Freebase-KG } \\
        & \multicolumn{2}{c|}{WebQSP} & \multicolumn{2}{c}{CWQ}  \\
        \midrule
        & Hit  & \#LLM  & Hit  & \#LLM  \\
        &(\%) & Calls & (\%) &Calls \\
        \midrule

        Vanilla LLM & 62.0 & 1 & 42.1 & 1 \\
        CoT LLM & 72.1 & 1 & 53.0 &1 \\
        KD-CoT~\citep{wang2023knowledge} & 68.6 & 2 &  55.7 & 4  \\
            StructGPT~\citep{jiang2023structgpt} & 72.6 & 4   &  54.2 & 4\\
            KB-BINDER~\citep{li2023binder} & 74.4 & 6 & --&  -- \\
            
            ToG~\citep{sun2024tog} & 82.6 & 11.2 & 67.6 & 14.3 \\
            
            EffiQA~\citep{dong-2025-effiqa} & 82.9 & 4.4 & 69.5 & 6.5 \\
            FiDeLiS~\citep{sui2024fidelis} & 84.1 & 10.7 & \textbf{71.4} & 15.2 \\
            \textbf{\byokg} (scoring)& 85.4 & 2 & 68.7 & 3  \\
             \textbf{\byokg} (agentic)& \textbf{87.1} & 4.5 & 71.1  & 6.3 \\

			\bottomrule
		\end{tabular}%
		\end{threeparttable}
}

\end{table}%

%% file: tables/ablation.tex
\begin{table}[tb]
	\centering
        	
        \caption{Ablation study on different linking methods (Claude-Sonnet-3.5 LLM).  }
        \label{tab:linker-results}%
	\resizebox{0.85\columnwidth}{!}{
    	
	\begin{threeparttable}
		\begin{tabular}{@{}l|cc@{}}
			\toprule
            
                & CWQ-IH & CronQuestions  \\
			  & Hit (\%) & Hit (\%) \\
        \midrule
        Vanilla LLM  & 54.2 & 19.2  \\
        \midrule
        Scoring-based Retrieval: & \\
        \; w/ Entity Linking only & 63.0 & 59.8 \\
        \; w/ \glink & 71.6 & 63.2 \\
        \midrule
        Agentic Retrieval: \\
        \; w/ Entity Linking only & 69.3 & 57.3 \\
        \; w/ \glink & 73.6 & 65.5 \\

			\bottomrule
		\end{tabular}%
		\end{threeparttable}
}

\end{table}%

%% file: tables/entity_linking.tex
\begin{table}[tb]
	\centering
	\caption{Comparison of different retrieval methods$^\dagger$ on CWQ.  `\#KG Tokens' denotes the median number of KG tokens$^\ddagger$ retrieved as context for the LLM.}
	\label{tab:retrieval}%
	\resizebox{\columnwidth}{!}{
	\begin{threeparttable}
		\begin{tabular}{@{}l|ccc@{}}
			\toprule
        & Recall@10 & \#KG Tokens$^\ddagger$ & \#Train Data\\ 
            \midrule
            RoG & 54.5 & 201  & 30.5K\\
            
            SubgraphRAG & 58.7 & 1,442 & 27.6K \\
            GNN-RAG& 64.1  &   114 & 27.6K  \\
            \midrule
            \byokg: \\
            \; - Scoring & 60.6 & 1,483 & 0\\
            \; - Agentic & 70.5 & 396  & 0 \\
			\bottomrule
		\end{tabular}%
        
		\begin{tablenotes}
        
        \item $^\dagger$RoG~\citep{luo2024rog}, SubgraphRAG~\citep{li2024subgraph}, and GNN-RAG~\citep{mavromatis2024gnn} are fine-tuned graph retrievers.
        \item $^\ddagger$We  count tokens via https://github.com/openai/tiktoken.

        \end{tablenotes}
		\end{threeparttable}
}

\end{table}%

%% file: tables/case.tex
\begin{table*}[h]
    \centering
    \caption{Case study on \byokg's iterative retrieval.}
    \vspace{-3mm}
    \label{tab:case}
    \resizebox{0.8\linewidth}{!}{%
    \begin{tabular}{@{}c|p{5in}@{}}
        \toprule
    Question & In what years did Stan Kasten's organization win the World Series? \\
    \midrule
    Answer & 1963 World Series | 1988 World Series | 1965 World Series | 1981 World Series | 1959 World Series \\
    \midrule
    \byokg & m.0\_yv0g3 -> organization.leadership.person -> Stan Kasten \\
     (1st iteration) & m.0\_yv0g3 -> organization.leadership.organization -> Los Angeles Dodgers \\
    & Stan Kasten -> business.board\_member.leader\_of > m.0\_yv0g3-> organization.leadership.organization -> Los Angeles Dodgers \\
    \midrule
    \byokg  & Los Angeles Dodgers -> sports.sports\_team.championships -> \textbf{1963 World Series |} \\
    (2nd iteration) & \textbf{1988 World Series | 1965 World Series | 1981 World Series | 1959 World Series} \\
  \bottomrule
    \end{tabular}}
    \vspace{-3mm}
\end{table*}

%% file: appendix.tex
\appendix

\section{Scoring-based Triplet Retrieval}
\label{app:scoring}

Alternatively to the the LLM-based traversal of \Secref{sec:agentic}, scoring-based retrieval performs top-$k$ triplet selection $\gT_q$ based on question-triplet semantic similarity, where $|\gT_q| = k$ and $k$ is a hyperparameter.

\subsection{Text-based Retrieval}
\label{app:text-retrieval}

Text-based retrieval converts KG triplets $\gT$ into natural language statements and retrieves the top-$k$ triplets based on their semantic similarity to the question. Since computing embeddings for all triplets in real-world KGs is computationally prohibitive, we adopt an efficient embedding decomposition approach~\citep{li2024subgraph}. Specifically, we decompose triplets into head, relation, and tail components, pre-compute embeddings for individual nodes and relations, and aggregate their similarity scores with the question. Formally, this approach is expressed as: \begin{align} 
\gT_q = \underset{(h,r,t) \in \gT}{\arg\text{top-}k}  \big( & \texttt{Embed}(q,h) + \nonumber \\ 
& \texttt{Embed}(q,r) + \nonumber\\ 
& \texttt{Embed}(q,t)\big), 
\label{eq:text-retrieval}
\end{align} 
where \texttt{Embed} computes cosine similarity using pretrained embedding models~\citep{chen2024bge}, implemented efficiently via FAISS~\citep{douze2024faiss}\footnote{https://huggingface.co/} .

\subsection{Graph Reranker}
\label{app:graph-reranker}
Given linked entities $\gE$ (\Secref{sec:linking}) and query $q$, the graph reranker module retrieves the top-$k$ most relevant triplets by reranking~\citep{gao2023retrieval} triplets within an $L$-hop neighborhood of $\gE$.

First, we collect triplets $\gT^{(L)}$ within an $L$-hop distance from $\gE$ as
\begin{equation}
   \gT^{(L)} = \bigcup_{e \in \gE} \{(h,r,t) \in \gG_{e}^{(L)}\}
\end{equation}
where $\gG_{e}^{(L)}$ denotes the $L$-hop subgraph starting from entity $e$, and $L$ is a hyperparameter that defaults to $L=2$.

Since the number of triplets $|\gT^{(L)}|$ can be excessively large in dense graphs, we implement a two-step pruning process: first filtering out triplets with irrelevant relations, then eliminating irrelevant triplets from the remaining set. We use a lightweight reranker model (\texttt{bge-reranker-base} cross-encoder~\citep{bge_embedding}) to maintain the most relevant relations
\begin{equation}
    \gR_{q} = \arg \text{top-}k_r\big(\texttt{Reranker}(q, \gR^{(L)})\big),
\end{equation}
which returns the top-$k_r$ most relevant relations to the question $q$, where $\gR^{(L)}$ is the relation set in $\gT^{(L)}$, $|\gR_{q}| = k_r$, and  $k_r = 20$ by default. 
Next, we collect triplets  of which relations are present in $\gR_{q}$ t9 $\gT_r^{(L)} = \{(h,r,t) \in \gT^{(L)}: r \in \gR_{q}\} $, and prune them again to $\gT_q^{(L)}$ by
\begin{equation}
    \gT_q^{(L)} = \arg \text{top-}k_t\big(\texttt{Reranker}(q, \gT_r^{(L)})\big),
\end{equation}
where $k_t = 100$ by default. Here, we transform the triplets to a natural language description for the reranker with a predefined template.
We use a more powerful reranker (\texttt{bge-reranker-v2-minicpm-layerwise} cross-encoder) to obtain the final top-$k$ most relevant triplets $\gT_q$ as
\begin{equation}
    \gT_{q} = \arg \text{top-}k \big(\texttt{Reranker2}(q, \gT_q^{(L)})\big),
\end{equation}
where $|\gT_{q}| = k$.

\input{tables/appendix/datasets}

\section{Dataset Statistics}
\label{app:data}

We provide the dataset statistics in Table~\ref{tab:datasets}, along with question examples from each benchmark.

\section{Other Evaluation Metrics}

\textbf{LLMaaJ}. We conducted preliminary evaluations using LLM-as-a-Judge (LLMaaJ) with Claude-Sonnet-3.5 on the WebQSP-IH and CWQ-IH datasets. The results show an alignment between the Hit@1 and LLMaaJ metrics for \byokg: 86.6\% Hit@1 vs. 86.8\% LLMaaJ on WebQSP-IH, and 73.6\% Hit@1 vs. 74.2\% LLMaaJ on CWQ-IH.

\textbf{F1}. F1 metric is suitable when question have multiple answers, common in KGQA tasks. \byokg is designed to support path-based retrieval and graph query execution, both of which naturally accommodate multiple valid answers. When \byokg successfully generates a correct path or Cypher query, it retrieves all associated anchor nodes, returning the full set of valid answers for the question.
We conducted a preliminary experiment on the WebQSP-IH dataset using path retrieval. On the subset of questions where the generated paths are executable, we found that the F1 score closely aligns with the Hit@1 metric (80.2\% Hit@1 vs. 78.8\% F1).

\section{Additional Experiments}
\label{app:exps}

\input{tables/appendix/refinement}

\input{tables/latency}

Table~\ref{tab:refine-results} presents the refinement effectiveness (\Secref{sec:refine}) of \byokg. In all cases presented, \byokg's  refinement improves the original graph retrieval methods considerably, highlighting the importance of iterative retrieval. To further analyze the impact of iteration count $T_R$, we ran additional experiments with 
$T_R=3$. KG-Linker includes a self-termination mechanism (Algorithm~\ref{alg:refine}), allowing the retrieval process to stop early once sufficient information has been gathered. For WebQSP-IH, KG-Linker terminates after an average of 1.1 iterations, achieving the same performance as with 
$T_R=2$. For CWQ-IH, it terminates after an average of 2.3 iterations, with a marginal performance gain (73.6\% for 
$T_R=2$ vs. 73.9\% for 
$T_R=3$). These findings suggest that the self-termination mechanism is effective in adapting to each query.

Table~\ref{tab:latency-results} presents a latency analysis, showing that while \byokg introduces moderate overhead $2.4\times$ relative to the baseline), it achieves superior performance than agentic retrieval ($1.6\times$).

\section{Case Studies}
\label{app:case}
\input{tables/appendix/cases}

We provide case studies on how \byokg improves retrieval in Table~\ref{tab:cases}. In CronQuestion, \byokg retrieves the correct answer while Agentic retrieval does not. In CWQ, \byokg finds the answers step-by-step (1st vs. 2nd iteration), while path retrieval also find relevant information. In MedQA, different components of \byokg retrieve information relevant to the correct answer. In Northwind, \byokg generates an executable cypehr query while graph-query does not.

\section{Prompts}
\subsection{\glink}
\label{app:linker-prompt}
\input{figures/appendix/entity_extraction}

\input{figures/appendix/path_extraction}

\input{figures/appendix/cypher_generation}

\input{figures/appendix/answer_generation}
The prompts per task employed in \glink (\Figref{fig:gconnector-prompt}) are presented in more details in \Figref{fig:linker-entity-prompt} (entity extraction), \Figref{fig:linker-path-prompt} (path extraction), \Figref{fig:linker-cypher-prompt} (graph query generation), and \Figref{fig:linker-answer-prompt} (answer generation).
\input{figures/appendix/relation_prompt}

\input{figures/appendix/entity_prompt}

\subsection{Agentic Traversal}
\label{app:agent-prompt}
The prompts used in agentic KG traversal are presented in \Figref{fig:agent-relation-prompt} (relation selection) and \Figref{fig:agent-entity-prompt} (entity selection).

%% file: tables/appendix/datasets.tex
\begin{table*}[tb]
	\centering
        	
        \caption{Summary of datasets.  }
        \label{tab:datasets}%
	\resizebox{\linewidth}{!}{
    	
	\begin{threeparttable}
		\begin{tabular}{@{}l|c|c|c|c@{}}
			\toprule
            
                & KG & \#Triplets  & \# QA & Type \\
        \midrule
        
        WebQSP  & Freebase &  16.3M & 1,638 & General Knowledge   \\
        WebQSP-IH  & Freebase & 1.5M & 500 & General Knowledge  \\
        CWQ & Freebase & 35.3M & 3,531 &  General Knowledge (multi-hop) \\
        CWQ-IH & Freebase & 1.5M & 500 &  General Knowledge (multi-hop) \\
        CronQuestions & Temp-Wikidata & 325K & 600 & Temporal Reasoning  \\
        MedQA & Disease DrugBank & 44.5K & 227& Medical Knowledge \\ 
        Northwind & Text2cypher KG & 3K & 178 & In-Domain Query \\
        \midrule
        \midrule
        & \multicolumn{4}{c}{Examples}  \\
        \midrule
        WebQSP  & \multicolumn{4}{c}{``\textit{In which state was the battle of Antietam fought?}''}  \\
        WebQSP-IH  & \multicolumn{4}{c}{``\textit{What language do Jamaican people speak?}''}  \\
        CWQ & \multicolumn{4}{c}{``\textit{Who was an actor in the film that Christopher Lee Foster was a crew member of as a kid?}''} \\
        CWQ-IH & \multicolumn{4}{c}{``\textit{"Lou Seal is the mascot for the team that last won the World Series when?}''} \\
        CronQuestions & \multicolumn{4}{c}{``\textit{Who preceded Diana Laidlaw as Minister for Transport?}''} \\
        MedQA & \multicolumn{4}{c}{``\textit{A 55-year-old woman has [condition]. What is best treatment?}''}\\ 
        Northwind &\multicolumn{4}{c}{ \textit{''Which categories have products with a unit price less than \$10?}''} \\
			\bottomrule
		\end{tabular}%
		\end{threeparttable}
}

\end{table*}%

%% file: tables/appendix/refinement.tex
\begin{table}[tb]
	\centering
        	
        \caption{Ablation study on retrieval refinement.  }
        \label{tab:refine-results}%
	\resizebox{0.85\columnwidth}{!}{
    	
	\begin{threeparttable}
		\begin{tabular}{@{}l|cc@{}}
			\toprule
            
                & CWQ-IH  \\
			  & Hit (\%)  \\
        Graph Reranker  & 63.0  \\
        \; +\byokg refinement  & 68.8 \\
        \midrule
        & WebQSP-IH  \\
        & Hit (\%)  \\
        Path Retrieval  & 75.2  \\
        \; +\byokg refinement  & 80.2 \\
        \midrule
        & NorthWind  \\
        & LLMaaJ (\%)  \\
        Graph-Query  & 55.3  \\
        \; + \byokg refinement  & 64.9 \\
			\bottomrule
		\end{tabular}%
		\end{threeparttable}
}

\end{table}%

%% file: tables/latency.tex
\begin{table}[tb]
	\centering
        	
        \caption{Comparison of average latency overhead across different methods on CWQ-IH. Results are normalized relative to the fastest method's per-query latency time, using Claude-Sonnet-3.5 as the backbone LLM.}
        \label{tab:latency-results}%
    \begin{threeparttable}
	\resizebox{0.9\columnwidth}{!}{

		\begin{tabular}{@{}l|cc@{}}
			\toprule
			  & Hit (\%) & Latency unit$^\dagger$ \\
              \midrule
              LLM+graph-query  & 54.4 & 1.0$\times$ \\
              Agentic Retrieval  & 69.3 & 1.6$\times$ \\
        \byokg & 73.6 & 2.4$\times$  \\

			\bottomrule
		\end{tabular}%
        }
        \begin{tablenotes}
        \small
        \item $^\dagger$We measured system latency on an EC2 g5.16xlarge instance, using Bedrock API for LLM queries. We acknowledge potential optimizations, including faster entity linking, deploying reranker and embedding models on higher-performance hardware, batch API calls, and prompt caching. 
         \end{tablenotes}
	\end{threeparttable}

\end{table}%

%% file: tables/appendix/cases.tex
\begin{table*}[]
    \centering
    \caption{Case studies on \byokg's benefits in KGQA.}
    \vspace{-3mm}
    \label{tab:cases}
    \resizebox{\linewidth}{!}{%
    \begin{tabular}{@{}c|p{5in}@{}}
        \toprule
        Question   (CronQuestions)          & When Francis Pym, Baron Pym had been the Member of the 48th Parliament of the United Kingdom, who had been the Minister of Finance of Norway? \\\midrule
        Answer               & Per Kleppe        \\  \midrule
        Agent  Retrieval    & Francis Pym, Baron Pym -> position held (1979 - 1983) -> Member of the 48th Parliament of the United Kingdom \\
         &   Francis Pym, Baron Pym -> position held (1974 - 1979) -> Member of the 47th Parliament of the United Kingdom \\ 
         &   [...] \\
            \midrule
        \byokg      & Member of the 48th Parliament of the United Kingdom -> position held (1979 - 1983) -> Francis Pym, Baron Pym \\
          &  Minister of Finance of Norway -> position held (1973 - 1979) -> \textbf{Per Kleppe} \\ 
          &  [...]        \\
            
            \\
        \midrule
        \midrule
    Question (CWQ) & In what years did Stan Kasten's organization win the World Series? \\
    \midrule
    Answer & 1963 World Series | 1988 World Series | 1965 World Series | 1981 World Series | 1959 World Series \\
    \midrule
    \byokg (1st iteration) & m.0\_yv0g3 -> organization.leadership.person -> Stan Kasten \\
    & m.0\_yv0g3 -> organization.leadership.organization -> Los Angeles Dodgers \\
    \midrule
    \byokg (2nd iteration) & Los Angeles Dodgers -> sports.sports\_team.championships -> \textbf{1963 World Series | 1988 World Series | 1965 World Series | 1981 World Series | 1959 World Series} \\
    \midrule
    \byokg (path retrieval) & Stan Kasten -> business.board\_member.leader\_of > m.0\_yv0g3-> organization.leadership.organization -> Los Angeles Dodgers \\
    \midrule
        \midrule
    Question (MedQA) & A 59-year-old overweight woman presents to the urgent care clinic with the complaint of severe abdominal pain for the past 2 hours. She also complains of a dull pain in her back with nausea and vomiting several times. Her pain has no relation with food. Her past medical history is significant for recurrent abdominal pain due to cholelithiasis. Her father died at the age of 60 with some form of abdominal cancer. Her temperature is 37C (98.6F), respirations are 15/min, pulse is 67/min, and blood pressure is 122/98 mm Hg. Physical exam is unremarkable. However, a CT scan of the abdomen shows a calcified mass near her gallbladder. Which of the following diagnoses should be excluded first in this patient \\
    \midrule
    Answer & Gallbladder cancer \\
    \midrule
    \byokg (agent) & Gallbladder cancer -> may cause -> Abdominal mass | Cholestatic jaundice | Liver metastases \\
    \midrule
    \byokg (path / query retrieval) & Aortic aneurysm, abdominal -> may cause > Abdominal mass-> may cause -> Gallbladder cancer \\
    \midrule
    \midrule

    Question (Northwind) & What is the average `unitPrice` of products that have been ordered in quantities greater than 10? \\
    \midrule
    Graph-Query & Query:  \textsc{MATCH (p:Product)<-[:PART\_OF]-(o:Order) WHERE o.quantity > 10 RETURN  avg(p.unitPrice) AS averageUnitPrice} \\
    & Result: \textsc{None} \\
    \midrule
    \byokg & Retrieval: \textsc{                 (638:  orderID: 10666, shipName: Richter Supermarkt) -> ORDERS  ->  (unitPrice: 123.79, productName: Thfcringer Rostbratwurst) }\\
     & Query: \textsc{MATCH (o:Order)-[r:ORDERS]->(p:Product) WHERE r.quantity > 10 RETURN avg(p.unitPrice) AS averageUnitPrice} \\
    & Result: \textsc{avgPrice: 26.0989786683906} \\
    
  \bottomrule
    \end{tabular}}
    \vspace{-3mm}
\end{table*}

%% file: figures/appendix/entity_extraction.tex
\begin{figure}
\begin{minipage}{1\columnwidth}
    \centering
\begin{tcolorbox}[title=\glink's Entity Extraction Prompt]
        \small
        \vspace{10pt}
        \textbf{Task: Entity Extraction} \\
        Extract all topic entities from the question (people, places, organizations, concepts, etc.) that will need to be matched in the graph database. \\
    - Include all entities that are directly mentioned and relevant to answering the question \\
    - Use exact names as they appear in the question \\
    - If an entity reference is vague or ambiguous, include both the mention from the question and potential aliases or full names that might match in the database \\
    - For organizations or entities with common abbreviations, include both full names and abbreviations when relevant \\
    - Format your response as follows, where entities are separated by newlines: \\

    <entities> \\
    entity1 \\
    entity1\_possible\_alias \\
    entity2 \\
    entity3 \\
    ... \\
    </entities>\\

    If no topic entities are present in the question, return empty tags:\\
    <entities>\\
    </entities>\\

    \textbf{Task: Relevant Entity Extraction}\\
    Extract all relevant entities from the graph context and question (people, places, organizations, concepts, etc.) that need to explore next for answering the question. \\
    Consider important entities only that are necessary for answering the question. Do not select entities, for which we already have all necessary information. \\
    - [Same instructions as before]

    - Format your response as follows, where entities are separated by newlines: \\

    <entities> \\
    next\_entity1 \\
    next\_entity1\_possible\_alias \\
    next\_entity2 \\
    next\_entity3 \\
    ... \\
    </entities> \\

    The entites should be sorted from the most important to the least important.  \\
    If we can answer the question directly based on the provided graph context, respond with: \\
    <entities> \\
    FINISH \\
    </entities> \\
    \end{tcolorbox}

\caption{The entity extraction prompt in \glink. We use the `relative' entity extraction prompt when provided with graph context.  }
\label{fig:linker-entity-prompt}
\end{minipage}
\end{figure}

%% file: figures/appendix/path_extraction.tex
\begin{figure}
\begin{minipage}{1\columnwidth}
    \centering
\begin{tcolorbox}[title=\glink's Path Generation Prompt]
        \small
        \vspace{10pt}
        \textbf{Task: Relationship Path Identification} \\
        Identify all relevant relationship paths that connect the entities and can be used to answer the question. \\
    - Only use relationships that are explicitly defined in the provided schema \\
    - Paths may be single relationships or combinations of multiple relationships  \\
    - Generate at least 3 different meaningful relationship paths when possible, focusing on diversity rather than redundancy \\
    - If graph context is provided, carefully analyze it to identify additional relevant relationship paths that might lead to the answer, especially focusing on paths to retrieve missing information that is not present in the context \\
    - Use the context to determine which paths are most likely to yield correct answers based on the information provided \\
    - Format your response as follows, where paths are separated by newlines: \\

    <paths> \\
    relation1 \\
    relation1 -> relation2 \\
    relation3 \\
    ... \\
    </paths> \\

    For multi-step paths, use the "->" delimiter between relationships. \\
    If the question does not require following any relationships, return empty tags: \\
    <paths> \\
    </paths> \\
    \end{tcolorbox}

\caption{The path generation prompt in \glink. }
\label{fig:linker-path-prompt}
\end{minipage}
\end{figure}

%% file: figures/appendix/cypher_generation.tex
\begin{figure}
\begin{minipage}{1\columnwidth}
    \centering
\begin{tcolorbox}[title=\glink's Graph Query Generation Prompt]
        \small
        \vspace{10pt}
     \textbf{Task: OpenCypher Query Generation} \\
    Construct a complete, executable OpenCypher statement that will retrieve the answer from a graph database. \\
    - Your query must only use node types, relationship types, and properties defined in the provided schema \\
    - Include appropriate MATCH patterns, WHERE clauses, and RETURN statements \\
    - Handle any filtering, aggregation, or sorting required by the question \\
    - If graph context is provided, ensure your query incorporates the relevant entities and relationships identified from the context\\
    
    - Format your response as follows: \\

    <opencypher> \\
    MATCH ... \\
    WHERE ... \\
    RETURN ... \\
    </opencypher> \\
    \end{tcolorbox}

\caption{The graph query generation prompt in \glink. }
\label{fig:linker-cypher-prompt}
\end{minipage}
\end{figure}

%% file: figures/appendix/answer_generation.tex
\begin{figure}
\begin{minipage}{1\columnwidth}
    \centering
\begin{tcolorbox}[title=\glink's Draft Answer Generation Prompt]
        \small
        \vspace{10pt}
        \textbf{Task: Question Answering} \\
        Answer the question using your existing knowledge base or the external information provided in the graph context (if provided).  \\
    - Provide only direct entity answers that specifically address the question \\
    - Each answer should be a distinct, well-defined entity (person, place, organization, concept, etc.) \\
    - List multiple answers if appropriate, with each answer on a separate line \\
    - Do not include explanations, reasoning, context, or commentary of any kind \\
    - Do not preface or conclude your answer with statements like "Based on my knowledge..." or "The answers are..." \\
    - If graph context is provided, prioritize answers that can be derived from the context over general knowledge\\
    - If you genuinely cannot determine the answer from the provided context or your knowledge base, you may return empty answer tags\\
    - Format your response exactly as follows, where answers are separated by newlines:\\

    <answers>\\
    answer\_entity1\\
    answer\_entity2\\
      ... \\
    </answers> \\

    If no clear answer can be determined, provide empty tags: \\
    <answers> \\
    </answers> \\
    \end{tcolorbox}

\caption{The answer generation prompt in \glink. }
\label{fig:linker-answer-prompt}
\end{minipage}
\end{figure}

%% file: figures/appendix/relation_prompt.tex
\begin{figure}
\begin{minipage}{1\columnwidth}
    \centering
\begin{tcolorbox}[title=Relation Selection Prompt]
        \small
        \vspace{10pt}
        \textbf{Task: Relation Selection} \\
Your task is to select the most appropriate relations based on their relevance to a given question.  \\

    Follow these steps: \\
    1. Read the provided <question>question</question> carefully. \\
    2. Analyze each relation in the <relation> list and determine its relevance to the question and relation. \\
    3. Respond by selecting the most relevant relations within <select>relations</select> tags. Be both frugal on your selection and consider completeness. \\
    4. The selected relations should be provided line-by-line. \\

    Example format:
    <question> \\
    Name the president of the country whose main spoken language was English in 1980? \\
    </question> \\

    <entity> \\
    English \\
    </entity> \\
 
    <relations> \\
    language.human\_language.main\_country \\
    language.human\_language.language\_family \\
    language.human\_language.iso\_639\_3\_code \\
    base.rosetta.languoid.parent \\
    language.human\_language.countries\_spoken\_in \\
    </relations> \\

    <selected> \\
    language.human\_language.main\_country \\
    language.human\_language.countries\_spoken\_in \\
    base.rosetta.languoid.parent \\
    </selected> \\

    Explanation: [short explanation (deprecated due to figure length)] \\

    Important Instructions: Always return at least one relation.
    Now it is your turn. 

    \vspace{5pt}
    <\textbf{question}> \\
    \{question\} \\
    </\textbf{question}> \\

    <\textbf{entity}> \\
    \{entity\}  \\
    </\textbf{entity}> \\
 
    <\textbf{relations}> \\
    \{relations\} \\
    </\textbf{relations}> \\

    Remember to parse your response in <selected></selected> tags:
    \end{tcolorbox}

\caption{The relation selection prompt template used in agentic traversal. }
\label{fig:agent-relation-prompt}
\end{minipage}
\end{figure}

%% file: figures/appendix/entity_prompt.tex
\begin{figure}
\begin{minipage}{1\columnwidth}
    \centering
\begin{tcolorbox}[title=Entity Selection Prompt]
        \small
        \vspace{10pt}
        \textbf{Task: Entity Selection} \\
        Given a question and the associated retrieved knowledge graph context (entity, relation, entity), you are asked to select the most important entities to explore in order to answer the question.
  Consider important entities only that are necessary for answering the question. Do not select entities, for which we already have all necessary information.

  - Format your response exactly as follows:
    <next-entities> \\
    relevant\_entity1 \\
    relevant\_entity2 \\
      ... \\
    </next-entities> \\
  
   The selected entities must be provided line-by-line.

  Example format:\\
    Question: Name the president of the country whose main spoken language was English in 1980? \\
    Graph Context: English -> countries\_spoken\_in -> England | USA \\

    <next-entities>\\
    England\\
    USA\\
    </next-entities>\\

  The entities should be sorted from the most important to the least important. 
  Important Instruction: If we can answer the question directly based on the provided graph context, respond with:\\
  <next-entities>\\
  FINISH\\
  </next-entities>\\

    \end{tcolorbox}

\caption{The entity selection prompt template used in agentic traversal. }
\label{fig:agent-entity-prompt}
\end{minipage}
\end{figure}

%% file: main_acl.bbl
\begin{thebibliography}{67}
\providecommand{\natexlab}[1]{#1}

\bibitem[{Agarwal et~al.(2023)Agarwal, Das, Khosla, and Gangadharaiah}]{agarwal2023bring}
Dhruv Agarwal, Rajarshi Das, Sopan Khosla, and Rashmi Gangadharaiah. 2023.
\newblock Bring your own kg: Self-supervised program synthesis for zero-shot kgqa.
\newblock \emph{arXiv preprint arXiv:2311.07850}.

\bibitem[{Anthropic(2024)}]{claude}
Anthropic. 2024.
\newblock \href {https://www-cdn.anthropic.com/de8ba9b01c9ab7cbabf5c33b80b7bbc618857627/Model_Card_Claude_3.pdf} {The claude 3 model family: Opus, sonnet, haiku}.

\bibitem[{Ao et~al.(2025)Ao, Yu, Wang, Deng, Guo, Pang, Wang, Chua, Zhang, and Cai}]{ao2025lightprof}
Tu~Ao, Yanhua Yu, Yuling Wang, Yang Deng, Zirui Guo, Liang Pang, Pinghui Wang, Tat-Seng Chua, Xiao Zhang, and Zhen Cai. 2025.
\newblock Lightprof: A lightweight reasoning framework for large language model on knowledge graph.
\newblock In \emph{Proceedings of the AAAI Conference on Artificial Intelligence}, volume~39, pages 23424--23432.

\bibitem[{Bollacker et~al.(2008)Bollacker, Evans, Paritosh, Sturge, and Taylor}]{bollacker2008freebase}
Kurt Bollacker, Colin Evans, Praveen Paritosh, Tim Sturge, and Jamie Taylor. 2008.
\newblock Freebase: a collaboratively created graph database for structuring human knowledge.
\newblock In \emph{Proceedings of the 2008 ACM SIGMOD international conference on Management of data}, pages 1247--1250.

\bibitem[{Bommasani et~al.(2021)Bommasani, Hudson, Adeli, Altman, Arora, von Arx, Bernstein, Bohg, Bosselut, Brunskill et~al.}]{bommasani2021opportunities}
Rishi Bommasani, Drew~A Hudson, Ehsan Adeli, Russ Altman, Simran Arora, Sydney von Arx, Michael~S Bernstein, Jeannette Bohg, Antoine Bosselut, Emma Brunskill, and 1 others. 2021.
\newblock On the opportunities and risks of foundation models.
\newblock \emph{arXiv preprint arXiv:2108.07258}.

\bibitem[{Brown et~al.(2020)Brown, Mann, Ryder, Subbiah, Kaplan, Dhariwal, Neelakantan, Shyam, Sastry, Askell et~al.}]{brown2020language}
Tom Brown, Benjamin Mann, Nick Ryder, Melanie Subbiah, Jared~D Kaplan, Prafulla Dhariwal, Arvind Neelakantan, Pranav Shyam, Girish Sastry, Amanda Askell, and 1 others. 2020.
\newblock Language models are few-shot learners.
\newblock \emph{Advances in neural information processing systems}, 33:1877--1901.

\bibitem[{Chen et~al.(2025)Chen, Guo, Yang, Chen, Chen, Liu, Shi, and Yang}]{chen2025pathrag}
Boyu Chen, Zirui Guo, Zidan Yang, Yuluo Chen, Junze Chen, Zhenghao Liu, Chuan Shi, and Cheng Yang. 2025.
\newblock Pathrag: Pruning graph-based retrieval augmented generation with relational paths.
\newblock \emph{arXiv preprint arXiv:2502.14902}.

\bibitem[{Chen et~al.(2024)Chen, Xiao, Zhang, Luo, Lian, and Liu}]{chen2024bge}
Jianlv Chen, Shitao Xiao, Peitian Zhang, Kun Luo, Defu Lian, and Zheng Liu. 2024.
\newblock Bge m3-embedding: Multi-lingual, multi-functionality, multi-granularity text embeddings through self-knowledge distillation.
\newblock \emph{arXiv preprint arXiv:2402.03216}.

\bibitem[{Dong et~al.(2025)Dong, Peng, Wang, Fu, Wang, Zhou, Shan, Zhu, and Chen}]{dong-2025-effiqa}
Zixuan Dong, Baoyun Peng, Yufei Wang, Jia Fu, Xiaodong Wang, Xin Zhou, Yongxue Shan, Kangchen Zhu, and Weiguo Chen. 2025.
\newblock {E}ffi{QA}: Efficient question-answering with strategic multi-model collaboration on knowledge graphs.
\newblock In \emph{Proceedings of the 31st International Conference on Computational Linguistics}.

\bibitem[{Douze et~al.(2024)Douze, Guzhva, Deng, Johnson, Szilvasy, Mazar{\'e}, Lomeli, Hosseini, and J{\'e}gou}]{douze2024faiss}
Matthijs Douze, Alexandr Guzhva, Chengqi Deng, Jeff Johnson, Gergely Szilvasy, Pierre-Emmanuel Mazar{\'e}, Maria Lomeli, Lucas Hosseini, and Herv{\'e} J{\'e}gou. 2024.
\newblock The faiss library.
\newblock \emph{arXiv preprint arXiv:2401.08281}.

\bibitem[{Edge et~al.(2024)Edge, Trinh, Cheng, Bradley, Chao, Mody, Truitt, and Larson}]{edge2024local}
Darren Edge, Ha~Trinh, Newman Cheng, Joshua Bradley, Alex Chao, Apurva Mody, Steven Truitt, and Jonathan Larson. 2024.
\newblock From local to global: A graph rag approach to query-focused summarization.
\newblock \emph{arXiv preprint arXiv:2404.16130}.

\bibitem[{Gao et~al.(2025)Gao, Cao, Bu, Zhu, Guan, and Yu}]{gao2025promoting}
Jianqi Gao, Jian Cao, Ranran Bu, Nengjun Zhu, Wei Guan, and Hang Yu. 2025.
\newblock Promoting knowledge base question answering by directing llms to generate task-relevant logical forms.
\newblock In \emph{Proceedings of the AAAI Conference on Artificial Intelligence}, volume~39, pages 23914--23922.

\bibitem[{Gao et~al.(2023)Gao, Xiong, Gao, Jia, Pan, Bi, Dai, Sun, and Wang}]{gao2023retrieval}
Yunfan Gao, Yun Xiong, Xinyu Gao, Kangxiang Jia, Jinliu Pan, Yuxi Bi, Yi~Dai, Jiawei Sun, and Haofen Wang. 2023.
\newblock Retrieval-augmented generation for large language models: A survey.
\newblock \emph{arXiv preprint arXiv:2312.10997}.

\bibitem[{Grattafiori et~al.(2024)Grattafiori, Dubey, Jauhri, Pandey, Kadian, Al-Dahle, Letman, Mathur, Schelten, Vaughan et~al.}]{grattafiori2024llama}
Aaron Grattafiori, Abhimanyu Dubey, Abhinav Jauhri, Abhinav Pandey, Abhishek Kadian, Ahmad Al-Dahle, Aiesha Letman, Akhil Mathur, Alan Schelten, Alex Vaughan, and 1 others. 2024.
\newblock The llama 3 herd of models.
\newblock \emph{arXiv preprint arXiv:2407.21783}.

\bibitem[{Gu et~al.(2023)Gu, Deng, and Su}]{gu-etal-2023-dont}
Yu~Gu, Xiang Deng, and Yu~Su. 2023.
\newblock Don{'}t generate, discriminate: A proposal for grounding language models to real-world environments.
\newblock In \emph{Proceedings of the 61st Annual Meeting of the Association for Computational Linguistics (Volume 1: Long Papers)}, Toronto, Canada. Association for Computational Linguistics.

\bibitem[{Gu and Su(2022)}]{gu2022arcaneqa}
Yu~Gu and Yu~Su. 2022.
\newblock Arcaneqa: Dynamic program induction and contextualized encoding for knowledge base question answering.
\newblock \emph{arXiv preprint arXiv:2204.08109}.

\bibitem[{Guo et~al.(2024)Guo, Xia, Yu, Ao, and Huang}]{guo2024lightrag}
Zirui Guo, Lianghao Xia, Yanhua Yu, Tu~Ao, and Chao Huang. 2024.
\newblock Lightrag: Simple and fast retrieval-augmented generation.

\bibitem[{Guti{\'e}rrez et~al.(2024)Guti{\'e}rrez, Shu, Gu, Yasunaga, and Su}]{gutierrez2024hipporag}
Bernal~Jim{\'e}nez Guti{\'e}rrez, Yiheng Shu, Yu~Gu, Michihiro Yasunaga, and Yu~Su. 2024.
\newblock Hipporag: Neurobiologically inspired long-term memory for large language models.
\newblock \emph{arXiv preprint arXiv:2405.14831}.

\bibitem[{He et~al.(2021)He, Lan, Jiang, Zhao, and Wen}]{he2021improving}
Gaole He, Yunshi Lan, Jing Jiang, Wayne~Xin Zhao, and Ji-Rong Wen. 2021.
\newblock Improving multi-hop knowledge base question answering by learning intermediate supervision signals.
\newblock In \emph{Proceedings of the 14th ACM international conference on web search and data mining}, pages 553--561.

\bibitem[{He et~al.(2024)He, Tian, Sun, Chawla, Laurent, LeCun, Bresson, and Hooi}]{he2024gretriever}
Xiaoxin He, Yijun Tian, Yifei Sun, Nitesh~V Chawla, Thomas Laurent, Yann LeCun, Xavier Bresson, and Bryan Hooi. 2024.
\newblock G-retriever: Retrieval-augmented generation for textual graph understanding and question answering.
\newblock \emph{arXiv preprint arXiv:2402.07630}.

\bibitem[{Jiang et~al.(2023{\natexlab{a}})Jiang, Zhou, Dong, Ye, Zhao, and Wen}]{jiang2023structgpt}
Jinhao Jiang, Kun Zhou, Zican Dong, Keming Ye, Wayne~Xin Zhao, and Ji-Rong Wen. 2023{\natexlab{a}}.
\newblock Structgpt: A general framework for large language model to reason over structured data.
\newblock \emph{arXiv preprint arXiv:2305.09645}.

\bibitem[{Jiang et~al.(2024)Jiang, Zhou, Zhao, Song, Zhu, Zhu, and Wen}]{jiang2024kg}
Jinhao Jiang, Kun Zhou, Wayne~Xin Zhao, Yang Song, Chen Zhu, Hengshu Zhu, and Ji-Rong Wen. 2024.
\newblock Kg-agent: An efficient autonomous agent framework for complex reasoning over knowledge graph.
\newblock \emph{arXiv preprint arXiv:2402.11163}.

\bibitem[{Jiang et~al.(2023{\natexlab{b}})Jiang, Zhou, Zhao, and Wen}]{jiang2023unikgqa}
Jinhao Jiang, Kun Zhou, Wayne~Xin Zhao, and Ji-Rong Wen. 2023{\natexlab{b}}.
\newblock Unikgqa: Unified retrieval and reasoning for solving multi-hop question answering over knowledge graph.
\newblock In \emph{International Conference on Learning Representations}.

\bibitem[{Jin et~al.(2021)Jin, Pan, Oufattole, Weng, Fang, and Szolovits}]{jin2021disease}
Di~Jin, Eileen Pan, Nassim Oufattole, Wei-Hung Weng, Hanyi Fang, and Peter Szolovits. 2021.
\newblock What disease does this patient have? a large-scale open domain question answering dataset from medical exams.
\newblock \emph{Applied Sciences}.

\bibitem[{Jo et~al.(2025)Jo, Choi, Kim, and Choi}]{jo2025r2}
Sumin Jo, Junseong Choi, Jiho Kim, and Edward Choi. 2025.
\newblock R2-kg: General-purpose dual-agent framework for reliable reasoning on knowledge graphs.
\newblock \emph{arXiv preprint arXiv:2502.12767}.

\bibitem[{Kim et~al.(2023)Kim, Kwon, Jo, and Choi}]{kim2023kggpt}
Jiho Kim, Yeonsu Kwon, Yohan Jo, and Edward Choi. 2023.
\newblock {KG}-{GPT}: A general framework for reasoning on knowledge graphs using large language models.
\newblock In \emph{Findings of the Association for Computational Linguistics: EMNLP 2023}.

\bibitem[{Lacroix et~al.(2020)Lacroix, Obozinski, and Usunier}]{lacroix2020tensor}
Timoth{\'e}e Lacroix, Guillaume Obozinski, and Nicolas Usunier. 2020.
\newblock Tensor decompositions for temporal knowledge base completion.
\newblock \emph{arXiv preprint arXiv:2004.04926}.

\bibitem[{Lan and Jiang(2020{\natexlab{a}})}]{lan-jiang-2020-query}
Yunshi Lan and Jing Jiang. 2020{\natexlab{a}}.
\newblock Query graph generation for answering multi-hop complex questions from knowledge bases.
\newblock In \emph{Proceedings of the 58th Annual Meeting of the Association for Computational Linguistics}, pages 969--974. Association for Computational Linguistics.

\bibitem[{Lan and Jiang(2020{\natexlab{b}})}]{lan2020query}
Yunshi Lan and Jing Jiang. 2020{\natexlab{b}}.
\newblock Query graph generation for answering multi-hop complex questions from knowledge bases.
\newblock Association for Computational Linguistics.

\bibitem[{Lee et~al.(2024)Lee, Zhu, Mavromatis, Han, Adeshina, Ioannidis, Rangwala, and Faloutsos}]{lee2024hybgrag}
Meng-Chieh Lee, Qi~Zhu, Costas Mavromatis, Zhen Han, Soji Adeshina, Vassilis~N Ioannidis, Huzefa Rangwala, and Christos Faloutsos. 2024.
\newblock Hybgrag: Hybrid retrieval-augmented generation on textual and relational knowledge bases.
\newblock \emph{arXiv preprint arXiv:2412.16311}.

\bibitem[{Lewis et~al.(2020)Lewis, Perez, Piktus, Petroni, Karpukhin, Goyal, K{\"u}ttler, Lewis, Yih, Rockt{\"a}schel et~al.}]{lewis2020retrieval}
Patrick Lewis, Ethan Perez, Aleksandra Piktus, Fabio Petroni, Vladimir Karpukhin, Naman Goyal, Heinrich K{\"u}ttler, Mike Lewis, Wen-tau Yih, Tim Rockt{\"a}schel, and 1 others. 2020.
\newblock Retrieval-augmented generation for knowledge-intensive nlp tasks.
\newblock \emph{Advances in Neural Information Processing Systems}, 33:9459--9474.

\bibitem[{Li et~al.(2020)Li, Min, Iyer, Mehdad, and Yih}]{li2020efficient}
Belinda~Z Li, Sewon Min, Srinivasan Iyer, Yashar Mehdad, and Wen-tau Yih. 2020.
\newblock Efficient one-pass end-to-end entity linking for questions.
\newblock \emph{arXiv preprint arXiv:2010.02413}.

\bibitem[{Li et~al.(2024{\natexlab{a}})Li, Zhang, Wu, Luo, Glass, and Meng}]{li2024decoding}
Kun Li, Tianhua Zhang, Xixin Wu, Hongyin Luo, James Glass, and Helen Meng. 2024{\natexlab{a}}.
\newblock Decoding on graphs: Faithful and sound reasoning on knowledge graphs through generation of well-formed chains.
\newblock \emph{arXiv preprint arXiv:2410.18415}.

\bibitem[{Li et~al.(2024{\natexlab{b}})Li, Miao, and Li}]{li2024subgraph}
Mufei Li, Siqi Miao, and Pan Li. 2024{\natexlab{b}}.
\newblock Simple is effective: The roles of graphs and large language models in knowledge-graph-based retrieval-augmented generation.
\newblock \emph{arXiv preprint arXiv:2410.20724}.

\bibitem[{Li et~al.(2023)Li, Ma, Zhuang, Gu, Su, and Chen}]{li2023binder}
Tianle Li, Xueguang Ma, Alex Zhuang, Yu~Gu, Yu~Su, and Wenhu Chen. 2023.
\newblock Few-shot in-context learning on knowledge base question answering.
\newblock In \emph{Proceedings of the 61st Annual Meeting of the Association for Computational Linguistics (Volume 1: Long Papers)}, pages 6966--6980. Association for Computational Linguistics.

\bibitem[{Liu et~al.(2023)Liu, Lin, Hewitt, Paranjape, Bevilacqua, Petroni, and Liang}]{liu2023lost}
Nelson~F Liu, Kevin Lin, John Hewitt, Ashwin Paranjape, Michele Bevilacqua, Fabio Petroni, and Percy Liang. 2023.
\newblock Lost in the middle: How language models use long contexts.
\newblock \emph{arXiv preprint arXiv:2307.03172}.

\bibitem[{Luo et~al.(2025{\natexlab{a}})Luo, Guo, Lin, Wu, Mu, Liu, Song, Zhu, Tuan et~al.}]{luo2025kbqao1}
Haoran Luo, Yikai Guo, Qika Lin, Xiaobao Wu, Xinyu Mu, Wenhao Liu, Meina Song, Yifan Zhu, Luu~Anh Tuan, and 1 others. 2025{\natexlab{a}}.
\newblock Kbqa-o1: Agentic knowledge base question answering with monte carlo tree search.
\newblock \emph{arXiv preprint arXiv:2501.18922}.

\bibitem[{Luo et~al.(2024{\natexlab{a}})Luo, Li, Haffari, and Pan}]{luo2024rog}
Linhao Luo, Yuan-Fang Li, Gholamreza Haffari, and Shirui Pan. 2024{\natexlab{a}}.
\newblock Reasoning on graphs: Faithful and interpretable large language model reasoning.
\newblock In \emph{International Conference on Learning Representations}.

\bibitem[{Luo et~al.(2024{\natexlab{b}})Luo, Zhao, Gong, Haffari, and Pan}]{luo2024graph}
Linhao Luo, Zicheng Zhao, Chen Gong, Gholamreza Haffari, and Shirui Pan. 2024{\natexlab{b}}.
\newblock Graph-constrained reasoning: Faithful reasoning on knowledge graphs with large language models.
\newblock \emph{arXiv preprint arXiv:2410.13080}.

\bibitem[{Luo et~al.(2025{\natexlab{b}})Luo, Zhao, Haffari, Phung, Gong, and Pan}]{luo2025gfm}
Linhao Luo, Zicheng Zhao, Gholamreza Haffari, Dinh Phung, Chen Gong, and Shirui Pan. 2025{\natexlab{b}}.
\newblock Gfm-rag: Graph foundation model for retrieval augmented generation.
\newblock \emph{arXiv preprint arXiv:2502.01113}.

\bibitem[{Mavromatis and Karypis(2022)}]{mavromatis2022rearev}
Costas Mavromatis and George Karypis. 2022.
\newblock \href {https://aclanthology.org/2022.findings-emnlp.181} {{R}ea{R}ev: Adaptive reasoning for question answering over knowledge graphs}.
\newblock In \emph{Findings of the Association for Computational Linguistics: EMNLP 2022}, pages 2447--2458, Abu Dhabi, United Arab Emirates. Association for Computational Linguistics.

\bibitem[{Mavromatis and Karypis(2024)}]{mavromatis2024gnn}
Costas Mavromatis and George Karypis. 2024.
\newblock G{NN}-{RAG}: Graph neural retrieval for large language model reasoning.
\newblock \emph{arXiv preprint arXiv:2405.20139}.

\bibitem[{Oliya et~al.(2021)Oliya, Saffari, Sen, and Ayoola}]{oliya2021end}
Armin Oliya, Amir Saffari, Priyanka Sen, and Tom Ayoola. 2021.
\newblock End-to-end entity resolution and question answering using differentiable knowledge graphs.
\newblock \emph{arXiv preprint arXiv:2109.05817}.

\bibitem[{Ozsoy et~al.(2024)Ozsoy, Messallem, Besga, and Minneci}]{ozsoy2024text2cypher}
Makbule~Gulcin Ozsoy, Leila Messallem, Jon Besga, and Gianandrea Minneci. 2024.
\newblock Text2cypher: Bridging natural language and graph databases.
\newblock \emph{arXiv preprint arXiv:2412.10064}.

\bibitem[{Pan et~al.(2024)Pan, Luo, Wang, Chen, Wang, and Wu}]{pan2024unifying}
Shirui Pan, Linhao Luo, Yufei Wang, Chen Chen, Jiapu Wang, and Xindong Wu. 2024.
\newblock Unifying large language models and knowledge graphs: A roadmap.
\newblock \emph{IEEE Transactions on Knowledge and Data Engineering}.

\bibitem[{Peng et~al.(2024)Peng, Zhu, Liu, Bo, Shi, Hong, Zhang, and Tang}]{peng2024graph}
Boci Peng, Yun Zhu, Yongchao Liu, Xiaohe Bo, Haizhou Shi, Chuntao Hong, Yan Zhang, and Siliang Tang. 2024.
\newblock Graph retrieval-augmented generation: A survey.
\newblock \emph{arXiv preprint arXiv:2408.08921}.

\bibitem[{Sarmah et~al.(2024)Sarmah, Mehta, Hall, Rao, Patel, and Pasquali}]{sarmah2024hybridrag}
Bhaskarjit Sarmah, Dhagash Mehta, Benika Hall, Rohan Rao, Sunil Patel, and Stefano Pasquali. 2024.
\newblock Hybridrag: Integrating knowledge graphs and vector retrieval augmented generation for efficient information extraction.
\newblock In \emph{Proceedings of the 5th ACM International Conference on AI in Finance}, pages 608--616.

\bibitem[{Saxena et~al.(2021)Saxena, Chakrabarti, and Talukdar}]{saxena2021question}
Apoorv Saxena, Soumen Chakrabarti, and Partha Talukdar. 2021.
\newblock Question answering over temporal knowledge graphs.
\newblock \emph{arXiv preprint arXiv:2106.01515}.

\bibitem[{Saxena et~al.(2020)Saxena, Tripathi, and Talukdar}]{saxena2020improving}
Apoorv Saxena, Aditay Tripathi, and Partha Talukdar. 2020.
\newblock Improving multi-hop question answering over knowledge graphs using knowledge base embeddings.
\newblock In \emph{Proceedings of the 58th Annual Meeting of the Association for Computational Linguistics}.

\bibitem[{Shi et~al.(2021)Shi, Cao, Hou, Li, and Zhang}]{shi2021transfernet}
Jiaxin Shi, Shulin Cao, Lei Hou, Juanzi Li, and Hanwang Zhang. 2021.
\newblock Transfernet: An effective and transparent framework for multi-hop question answering over relation graph.
\newblock \emph{arXiv preprint arXiv:2104.07302}.

\bibitem[{Shu et~al.(2022)Shu, Yu, Li, Karlsson, Ma, Qu, and Lin}]{shu2022tiara}
Yiheng Shu, Zhiwei Yu, Yuhan Li, B{\"o}rje~F Karlsson, Tingting Ma, Yuzhong Qu, and Chin-Yew Lin. 2022.
\newblock Tiara: Multi-grained retrieval for robust question answering over large knowledge bases.
\newblock \emph{arXiv preprint arXiv:2210.12925}.

\bibitem[{Sui et~al.(2024)Sui, He, Liu, He, Wang, and Hooi}]{sui2024fidelis}
Yuan Sui, Yufei He, Nian Liu, Xiaoxin He, Kun Wang, and Bryan Hooi. 2024.
\newblock Fidelis: Faithful reasoning in large language model for knowledge graph question answering.
\newblock \emph{arXiv preprint arXiv:2405.13873}.

\bibitem[{Sun et~al.(2018)Sun, Dhingra, Zaheer, Mazaitis, Salakhutdinov, and Cohen}]{sun-etal-2018-open}
Haitian Sun, Bhuwan Dhingra, Manzil Zaheer, Kathryn Mazaitis, Ruslan Salakhutdinov, and William Cohen. 2018.
\newblock Open domain question answering using early fusion of knowledge bases and text.
\newblock In \emph{Proceedings of the 2018 Conference on Empirical Methods in Natural Language Processing}, pages 4231--4242. Association for Computational Linguistics.

\bibitem[{Sun et~al.(2024)Sun, Xu, Tang, Wang, Lin, Gong, Shum, and Guo}]{sun2024tog}
Jiashuo Sun, Chengjin Xu, Lumingyuan Tang, Saizhuo Wang, Chen Lin, Yeyun Gong, Heung-Yeung Shum, and Jian Guo. 2024.
\newblock Think-on-graph: Deep and responsible reasoning of large language model with knowledge graph.
\newblock In \emph{International Conference on Learning Representations}.

\bibitem[{Sun et~al.(2020)Sun, Zhang, Cheng, and Qu}]{sun2020sparqa}
Yawei Sun, Lingling Zhang, Gong Cheng, and Yuzhong Qu. 2020.
\newblock Sparqa: skeleton-based semantic parsing for complex questions over knowledge bases.
\newblock In \emph{Proceedings of the AAAI conference on artificial intelligence}, volume~34, pages 8952--8959.

\bibitem[{Talmor and Berant(2018)}]{talmor2018web}
Alon Talmor and Jonathan Berant. 2018.
\newblock The web as a knowledge-base for answering complex questions.
\newblock In \emph{Proceedings of the 2018 Conference of the North {A}merican Chapter of the Association for Computational Linguistics}.

\bibitem[{Vrande{\v{c}}i{\'c} and Kr{\"o}tzsch(2014)}]{vrandevcic2014wikidata}
Denny Vrande{\v{c}}i{\'c} and Markus Kr{\"o}tzsch. 2014.
\newblock Wikidata: a free collaborative knowledgebase.
\newblock \emph{Communications of the ACM}, 57(10):78--85.

\bibitem[{Wang et~al.(2023)Wang, Duan, Wang, Li, Xian, Yin, Rong, and Xiong}]{wang2023knowledge}
Keheng Wang, Feiyu Duan, Sirui Wang, Peiguang Li, Yunsen Xian, Chuantao Yin, Wenge Rong, and Zhang Xiong. 2023.
\newblock Knowledge-driven cot: Exploring faithful reasoning in llms for knowledge-intensive question answering.
\newblock \emph{arXiv preprint arXiv:2308.13259}.

\bibitem[{Wishart et~al.(2018)Wishart, Feunang, Guo, Lo, Marcu, Grant, Sajed, Johnson, Li, Sayeeda et~al.}]{wishart2018drugbank}
David~S Wishart, Yannick~D Feunang, An~C Guo, Elvis~J Lo, Ana Marcu, Jason~R Grant, Tanvir Sajed, Daniel Johnson, Carin Li, Zinat Sayeeda, and 1 others. 2018.
\newblock Drugbank 5.0: a major update to the drugbank database for 2018.
\newblock \emph{Nucleic acids research}, 46(D1):D1074--D1082.

\bibitem[{Wu et~al.(2024)Wu, Zhao, Yasunaga, Huang, Cao, Huang, Ioannidis, Subbian, Zou, and Leskovec}]{wu2024stark}
Shirley Wu, Shiyu Zhao, Michihiro Yasunaga, Kexin Huang, Kaidi Cao, Qian Huang, Vassilis Ioannidis, Karthik Subbian, James~Y Zou, and Jure Leskovec. 2024.
\newblock Stark: Benchmarking llm retrieval on textual and relational knowledge bases.
\newblock \emph{Advances in Neural Information Processing Systems}, 37:127129--127153.

\bibitem[{Xiao et~al.(2023)Xiao, Liu, Zhang, and Muennighoff}]{bge_embedding}
Shitao Xiao, Zheng Liu, Peitian Zhang, and Niklas Muennighoff. 2023.
\newblock \href {https://arxiv.org/abs/2309.07597} {C-pack: Packaged resources to advance general chinese embedding}.
\newblock \emph{Preprint}, arXiv:2309.07597.

\bibitem[{Yadav and Bethard(2019)}]{yadav2019survey}
Vikas Yadav and Steven Bethard. 2019.
\newblock A survey on recent advances in named entity recognition from deep learning models.
\newblock \emph{arXiv preprint arXiv:1910.11470}.

\bibitem[{Yasunaga et~al.(2021)Yasunaga, Ren, Bosselut, Liang, and Leskovec}]{yasunaga2021qagnn}
Michihiro Yasunaga, Hongyu Ren, Antoine Bosselut, Percy Liang, and Jure Leskovec. 2021.
\newblock Qa-gnn: Reasoning with language models and knowledge graphs for question answering.
\newblock In \emph{North American Chapter of the Association for Computational Linguistics (NAACL)}.

\bibitem[{Ye et~al.(2022)Ye, Yavuz, Hashimoto, Zhou, and Xiong}]{ye-etal-2022-rng}
Xi~Ye, Semih Yavuz, Kazuma Hashimoto, Yingbo Zhou, and Caiming Xiong. 2022.
\newblock {RNG}-{KBQA}: Generation augmented iterative ranking for knowledge base question answering.
\newblock In \emph{Proceedings of the 60th Annual Meeting of the Association for Computational Linguistics (Volume 1: Long Papers)}, pages 6032--6043. Association for Computational Linguistics.

\bibitem[{Yih et~al.(2015)Yih, Chang, He, and Gao}]{yih2015semantic}
Scott Wen-tau Yih, Ming-Wei Chang, Xiaodong He, and Jianfeng Gao. 2015.
\newblock Semantic parsing via staged query graph generation: Question answering with knowledge base.
\newblock In \emph{Proceedings of the Joint Conference of the 53rd Annual Meeting of the ACL and the 7th International Joint Conference on Natural Language Processing of the AFNLP}.

\bibitem[{Yu et~al.(2022)Yu, Zhang, Ng, Zhu, Li, Wang, Hu, Wang, Wang, and Xiang}]{yu2022decaf}
Donghan Yu, Sheng Zhang, Patrick Ng, Henghui Zhu, Alexander~Hanbo Li, Jun Wang, Yiqun Hu, William Wang, Zhiguo Wang, and Bing Xiang. 2022.
\newblock Decaf: Joint decoding of answers and logical forms for question answering over knowledge bases.
\newblock \emph{arXiv preprint arXiv:2210.00063}.

\bibitem[{Zhang et~al.(2025)Zhang, Zhou, and Yang}]{zhang2025learning}
Han Zhang, Langshi Zhou, and Hanfang Yang. 2025.
\newblock Learning to retrieve and reason on knowledge graph through active self-reflection.
\newblock \emph{arXiv preprint arXiv:2502.14932}.

\end{thebibliography}
